\documentclass[conference]{IEEEtran}
\IEEEoverridecommandlockouts
\usepackage{cite}
\usepackage{amsmath,amssymb,amsfonts}
\usepackage{algorithmic}
\usepackage{graphicx}
\usepackage{hyperref}
\usepackage{textcomp}
\usepackage[table]{xcolor}
\usepackage{tabularx}
\usepackage{colortbl}
\usepackage{booktabs}
\usepackage{multirow}
\usepackage{silence}
\usepackage{subcaption}
\usepackage{xcolor}
\def\BibTeX{{\rm B\kern-.05em{\sc i\kern-.025em b}\kern-.08em
    T\kern-.1667em\lower.7ex\hbox{E}\kern-.125emX}}
\begin{document}

\title{Detecting Hate and Inflammatory Content in Bengali Memes\thanks{The dataset is publicly available at \url{https://doi.org/10.17632/9vg79v65nr.1}}: A New Multimodal Dataset and Co-Attention Framework\\}

\author{\IEEEauthorblockN{Rakib Ullah}
\IEEEauthorblockA{\textit{Computer Science And Engineering} \\
\textit{Sylhet Engineering College}\\
Sylhet, Bangladesh \\
rakib14@sec.ac.bd}
\and
\IEEEauthorblockN{Mominul islam}
\IEEEauthorblockA{\textit{Computer Science And Engineering} \\
\textit{Daffodil International University}\\
Dhaka, Bangladesh \\
mominul11992@gmail.com}
\and
\IEEEauthorblockN{Md Sanjid Hossain}
\IEEEauthorblockA{\textit{Computer Science And Engineering} \\
\textit{Daffodil International University}\\
Dhaka, Bangladesh \\
sanjid15-11888@diu.edu.bd}
\and
\IEEEauthorblockN{Md Ismail Hossain}
\IEEEauthorblockA{\textit{Computer Science And Engineering} \\
\textit{Daffodil International University}\\
Dhaka, Bangladesh \\
ismail.hossain12012@gmail.com}

}

\maketitle

\begin{abstract}
Internet memes have become a dominant form of expression on social media, including within the Bengali-speaking community. While often humorous, memes can also be exploited to spread offensive, harmful, and inflammatory content targeting individuals and groups. Detecting this type of content is exceptionally challenging due to its satirical, subtle, and culturally specific nature. This problem is magnified for low-resource languages like Bengali, as existing research predominantly focuses on high-resource languages. To address this critical research gap, we introduce Bn-HIB (Bangla Hate Inflammatory Benign), a novel dataset containing 3,247 manually annotated Bengali memes categorized as Benign, Hate, or Inflammatory. Significantly, Bn-HIB is the first dataset to distinguish inflammatory content from direct hate speech in Bengali memes. Furthermore, we propose the MCFM (Multi-Modal Co-Attention Fusion Model), a simple yet effective architecture that mutually analyzes both the visual and textual elements of a meme. MCFM employs a co-attention mechanism to identify and fuse the most critical features from each modality, leading to a more accurate classification. Our experiments show that MCFM significantly outperforms several state-of-the-art models on the Bn-HIB dataset, demonstrating its effectiveness in this nuanced task.
\\ To facilitate reproducibility and future research, the Bn-HIB dataset has been made publicly available through Mendeley Data. \textcolor{red}{\textbf{Warning:} This work contains material that may be disturbing to some audience members. Viewer discretion is advised.}

\end{abstract}


\section{Introduction}
Image-text memes have become a powerful communication medium but also a vehicle for hate speech, misinformation, and polarization. Their multi-modal and culturally nuanced nature makes harmful intent detection difficult, especially in low-resource languages like Bengali, where code-mixing and symbolic visuals challenge existing English-centric moderation systems.
As memes increasingly shape social interactions, research interest in meme analysis has grown substantially. Prior works explored various communicative dimensions such as sentiment and emotion recognition~\cite{mishra2023memotion3datasetsentiment}, sarcasm detection~\cite{10.1007/978-3-031-28244-7_7}, and offensive content identification~\cite{inproceedings}. The rise of toxic memes has further driven studies on hate~\cite{DBLP:journals/corr/abs-2005-04790, KAPIL2025100133, el-sayed-nasr-2024-aast-nlp}, offensiveness~\cite{SHANG2021102664, KUMARI2025101781}, and harmful intent~\cite{pramanick-etal-2021-momenta-multimodal}. However, these efforts have largely focused on high-resource languages (e.g., English, Hindi), leaving low-resource languages underexplored. Only a few works, such as those on Tamil, Malayan, and other regional languages~\cite{suryawanshi-chakravarthi-2021-findings, kumari2023, manukonda-kodali-2025-bytesizedllm-dravidianlangtech-2025}, have addressed objectionable memes, revealing a critical gap in low-resource meme moderation research.
Bengali memes have recently gained wide popularity, shaping public opinion but often promoting negativity and violence. Prior studies have explored abusive~\cite{das2023banglaabusememedatasetbengaliabusive}, hateful~\cite{10.1007/978-3-031-33231-9_21,hossain-etal-2022-mute,hossain2024,debnath-etal-2025-exmute}, and aggressive~\cite{ahsan2024} Bengali memes. However, these works are limited to binary or topic-specific hate detection (e.g., religion, gender, politics etc.).

\begin{table}
\centering
\scriptsize
\caption{Annotation Inclusion And Exclusion Criteria}
\label{tab:annotation-guidelines}
\renewcommand{\arraystretch}{0.9}
\setlength{\tabcolsep}{2pt}
\begin{tabularx}{\columnwidth}{|p{2.1cm}|X|X|}
\hline
\textbf{Class} & \textbf{Inclusion Criteria} & \textbf{Exclusion Criteria} \\
\hline
\textbf{Inflammatory} &
-- Controversial statements or imagery; norm-challenging discourse; inflammatory tone or humor; boundary-pushing content. &
-- Accidental controversy; mild disagreement; primarily educational focus. \\
\hline
\textbf{Hate} &
-- Specific names or groups; group-identifiable traits; pronouns with clear referents; direct or indirect addressing. &
-- Generic mentions; self-referential content. \\
\hline
\textbf{Benign} &
-- Positive or neutral tone; everyday relatable humor; inclusive, non-divisive content. &
-- Any provocative or targeting cue; implicit controversial element. \\
\hline
\end{tabularx}
\end{table}

\begin{figure}
    \centering
    \begin{subfigure}{0.48\columnwidth}
        \centering
        \includegraphics[width=\linewidth]{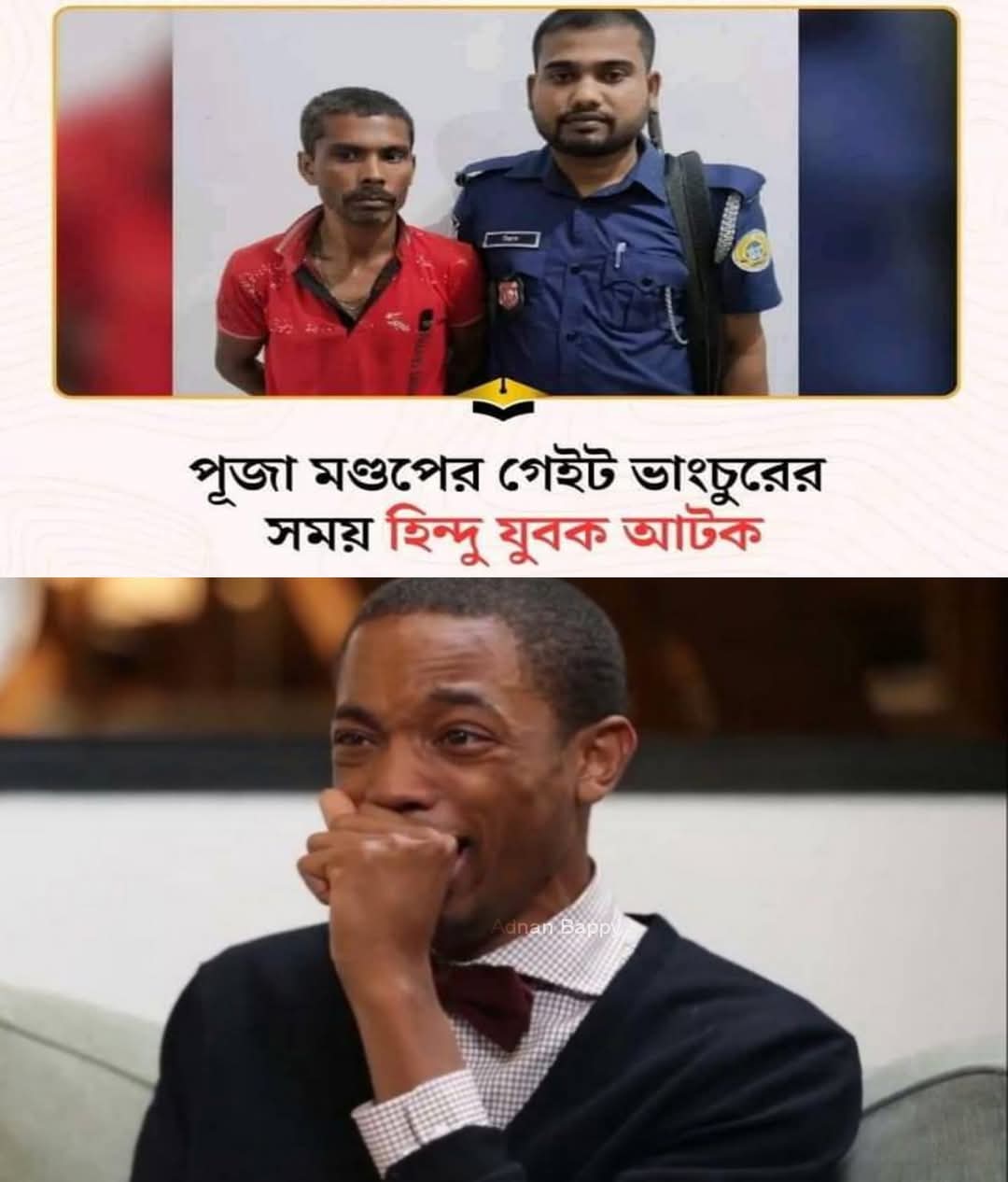}
        \textbf{English Translation}:Hindu man arrested while vandalizing a Hindu temple gate
        \caption{}
        \label{fig:inflamatory meme}
    \end{subfigure}
    \hfill
    \begin{subfigure}{0.48\columnwidth}
        \centering
        \includegraphics[width=\linewidth]{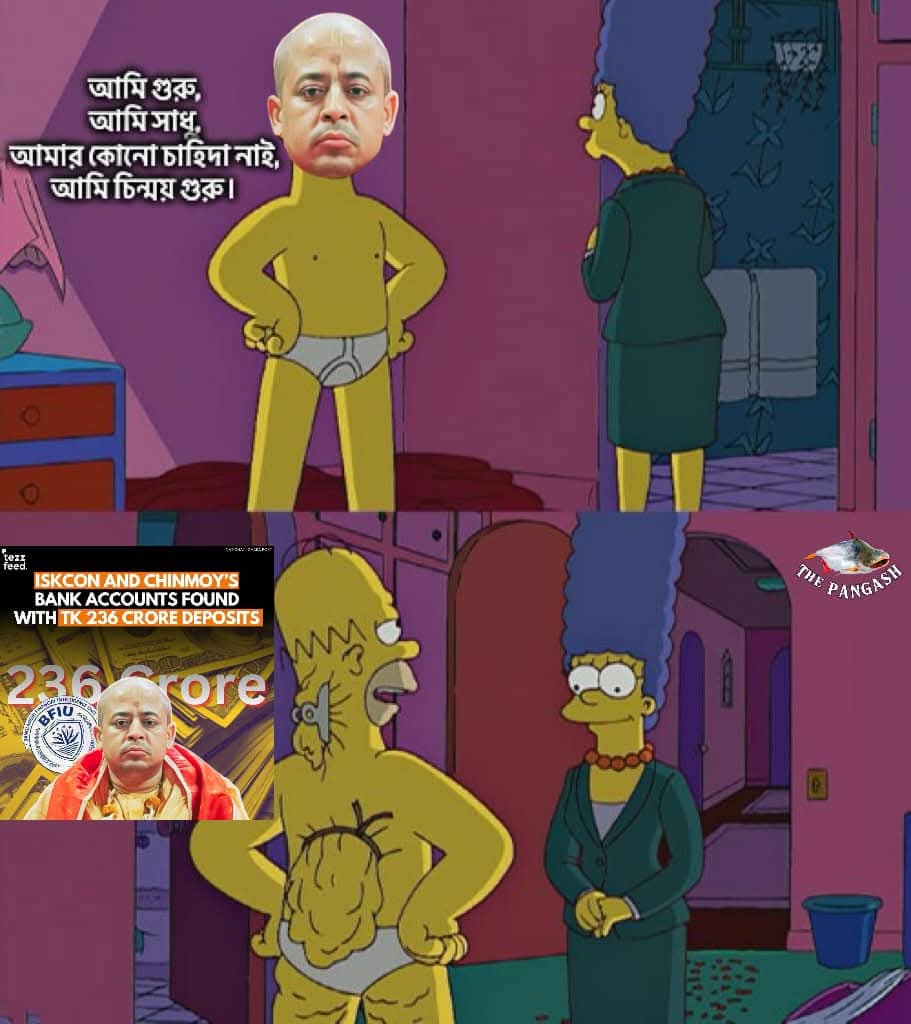}
        \textbf{English Translation}:I am a guru, I am a sage, I have no desires, I am Chinmoy guru.\\
        Iskcon and Chinmoy's Bank account found with TK 236 crore deposits
        \caption{}
        \label{fig:hateful meme}
    \end{subfigure}
    \caption{Meme (a) portrays Hindus as hypocrites for allegedly destroying their own temples and blaming Muslims, while Meme (b) specifically targets an organization and individual, making it more directed but less representative of the broader Hindu population.}
    \label{fig:hate vs inflamatory}
\end{figure}

Research in low-resource settings has largely overlooked the distinction between hateful and inflammatory memes. While not all hateful memes are designed to provoke communal tensions or incite violence, inflammatory memes subtly manipulate public perception, fostering generalized negative attitudes toward specific groups. Figure \ref{fig:hate vs inflamatory} illustrates this distinction: although Fig. \ref{fig:inflamatory meme} may appear hateful, a detailed analysis of its textual and visual context classifies it as inflammatory due to its implicit manipulative persuasive intent. Previous systems fail to differentiate such subtle cases, highlighting the need for dedicated approaches.

Motivated by this gap, we introduce a novel corpus of Bengali memes, differentiating inflammatory, hateful, and benign content. On the technical front, state-of-the-art multimodal systems struggle with meme analysis, particularly when subtlety or implicit meaning is involved. Memes often exhibit weak alignment between text and visuals and contain significant noise compared to structured multi-modal data. To address these challenges, we propose a Multi-Modal Co-Attention Fusion Model (MCFM).

\begin{itemize}
    \item We develop a novel multi-modal
    dataset Bn-HIB consisting of 3,247  Bengali memes labeled as Hate, Benign and  Inflammatory. 
    \item We propose MCFM, a simple yet effective multimodal fusion model that leverages both combined and modality-specific features to capture subtle elements.
    \item Finally, We conduct extensive experiments showing that MCFM outperforms fifteen state-of-the-art unimodal and multimodal baselines across all evaluation metrics.
\end{itemize}

\section{Related Work}
This section reviews prior studies on multimodal hate speech detection datasets in both high-resource and low-resource language settings.\\
\\
\textbf{Multi-Modal Hate Speech Detection Datasets In High-Resource Settings:}
Several studies have explored multimodal information for hate speech and offensive content detection in recent years. 
\cite{suryawanshi2020a} developed a multimodal dataset for offensive meme detection. 
Similarly, \cite{kiela2020hateful} and \cite{gomez2020exploring} introduced multimodal datasets focused on detecting hateful content from online memes. 
\cite{pramanick-etal-2021-momenta-multimodal} proposed the MOMENTA dataset for harmful meme detection, incorporating both textual and visual modalities. 
In addition, two multimodal datasets have been developed for the Hindi language: one by \cite{kumari2023} and another by \cite{rajput2022hatenotdetectinghate}, both aimed at identifying offensive and hateful memes in a high-resource context.

\begin{figure}
    \centering
    \includegraphics[width=\linewidth]{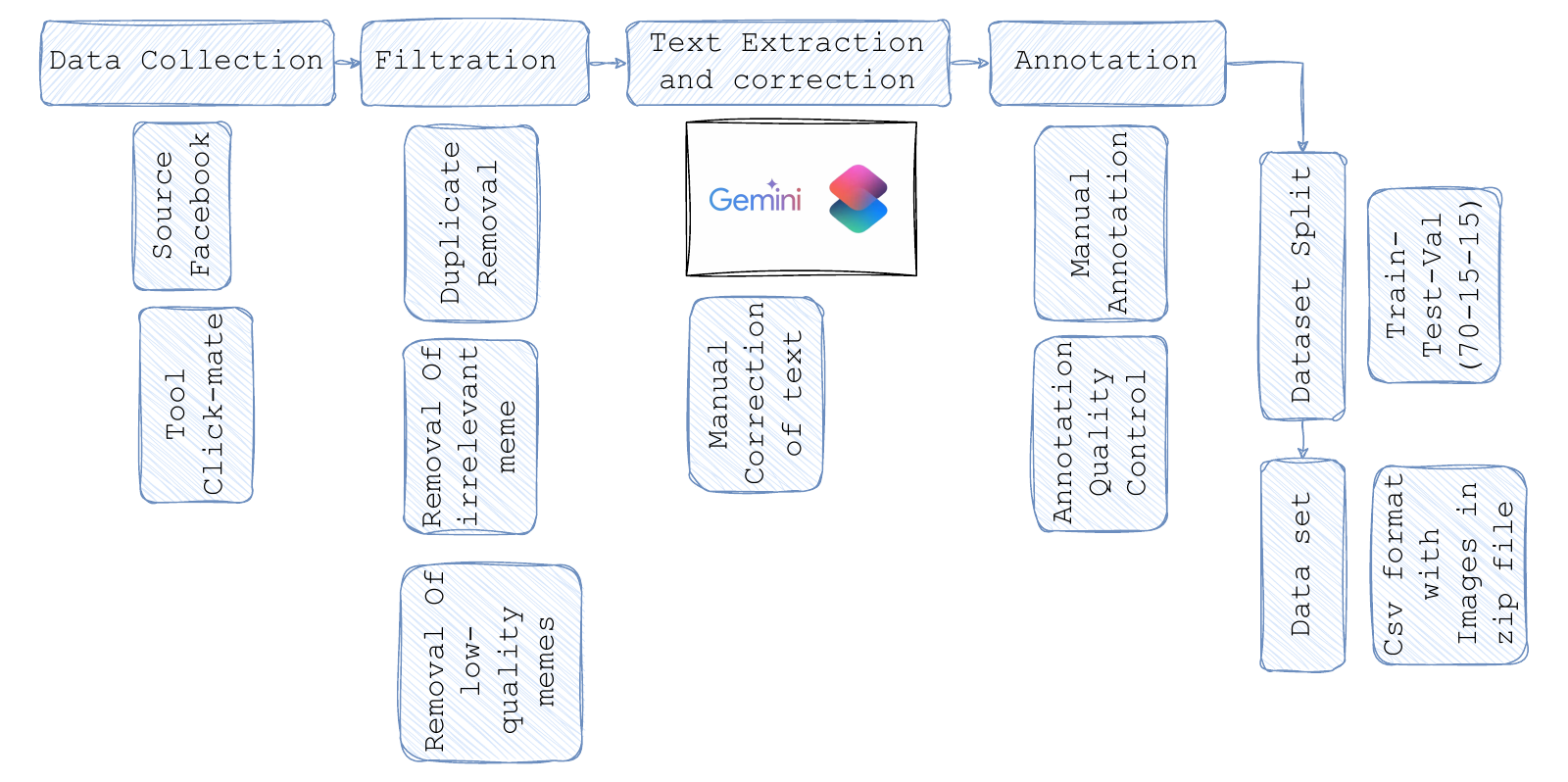}
    \caption{Data Collection and Pre-Processing Overview}
    \label{fig:Data Pre-Processing Diagram}
\end{figure}

\textbf{Multi-Modal Hate Speech Detection Datasets In Low Resource Settings:}
Research on multimodal hate speech detection in low-resource languages such as Bengali has gained attention more recently. 
\cite{karim2022}, \cite{hossain2024}, \cite{hossain-etal-2022-mute}, and \cite{debnath-etal-2025-exmute} introduced multimodal hate speech datasets addressing the Bangla language. 
\cite{das2023banglaabusememedatasetbengaliabusive} developed a dataset for abusive meme detection in Bengali, while \cite{ahsan2024} focused on aggression detection in similar multimodal contexts. 
Researchers have employed various fusion techniques to combine visual and textual modalities, including early and late fusion methods, and the use of pretrained vision-language models such as CLIP \cite{das-mukherjee-2023-banglaabusememe,ahsan2024,hossain2024}, BLIP \cite{ahsan2024}, ALBEF \cite{ahsan2024,hossain2024}, and MMBT \cite{hossain2024}. 
Some works have also introduced custom architectures \cite{hossain2024,ahsan2024} tailored to their specific tasks.
\\
\textbf{Differences From Existing Research:} While significant progress has been made in multimodal hate speech and offensive content detection, there remains a notable gap in the study of inflammatory meme detection, particularly for low-resource languages such as Bengali. 
To the best of our knowledge, no prior work distinguishes hate from inflammatory content within this domain. 
Moreover, most existing datasets focus on binary classifications (e.g., hateful vs. non-hateful), limiting their capacity to capture nuanced or targeted offensive content—such as political, gendered, or religious themes—which are prevalent and rapidly evolving across social media platforms. Most of the existing datasets are entirely in Bengali \cite{hossain-etal-2022-mute,das-mukherjee-2023-banglaabusememe,karim2022,das2023banglaabusememedatasetbengaliabusive,ahsan2024}. Only one study focuses on Bengali and code-mixed Bengali \cite{hossain2024}, while another includes code-mixed, code-switched, and Bengali Memes \cite{debnath-etal-2025-exmute}.Furthermore, No previous study has publicly disclosed the specific sources of their meme collections.

In light of these identified
gaps, our work differs from the existing works in
three significant ways: 
(i) we develop a dataset distinguishing \textit{hate}, \textit{inflammatory}, and \textit{benign} memes in Bengali, Bengali-English code-mixed, and code-switched contexts; 
(ii) we provide an annotation guideline to support dataset creation for other low-resource languages.(iii) we identified and shared Public Bengali Facebook meme pages and groups with front-page screenshots for efficient meme sourcing for future researchers.

\section{Dataset}
Description of Classes:
\begin{itemize}
\item \textbf{Hate Memes (HM):} Explicitly hostile content targeting individuals or groups through derogatory language, visuals, or stereotypes, aiming to incite hatred.
\item \textbf{Inflammatory Memes (IM):} Implicitly Inflammatory content using sarcasm, irony, or coded cues to incite polarization or unrest without overt hate speech.
\item \textbf{Benign Memes (BM):} Non-hostile, humorous, or cultural content intended for entertainment or social bonding without divisive intent.
\end{itemize}

Data was collected from 25 public Facebook groups/pages known for high meme activity using manual curation followed by the Click-mate\cite{click-mate} macro tool for scalable extraction. Only publicly available content was used, yielding approximately 5,000 raw memes. After filtering for multi-modality and quality as\cite{hossain-etal-2022-mute}, 3,247 memes remained. Text embedded in images was extracted using the Gemini API\cite{GeminiApi}, manually verified for Bengali script accuracy, and paired with corresponding images.  

\begin{figure}
    \centering
    \includegraphics[width=\linewidth]{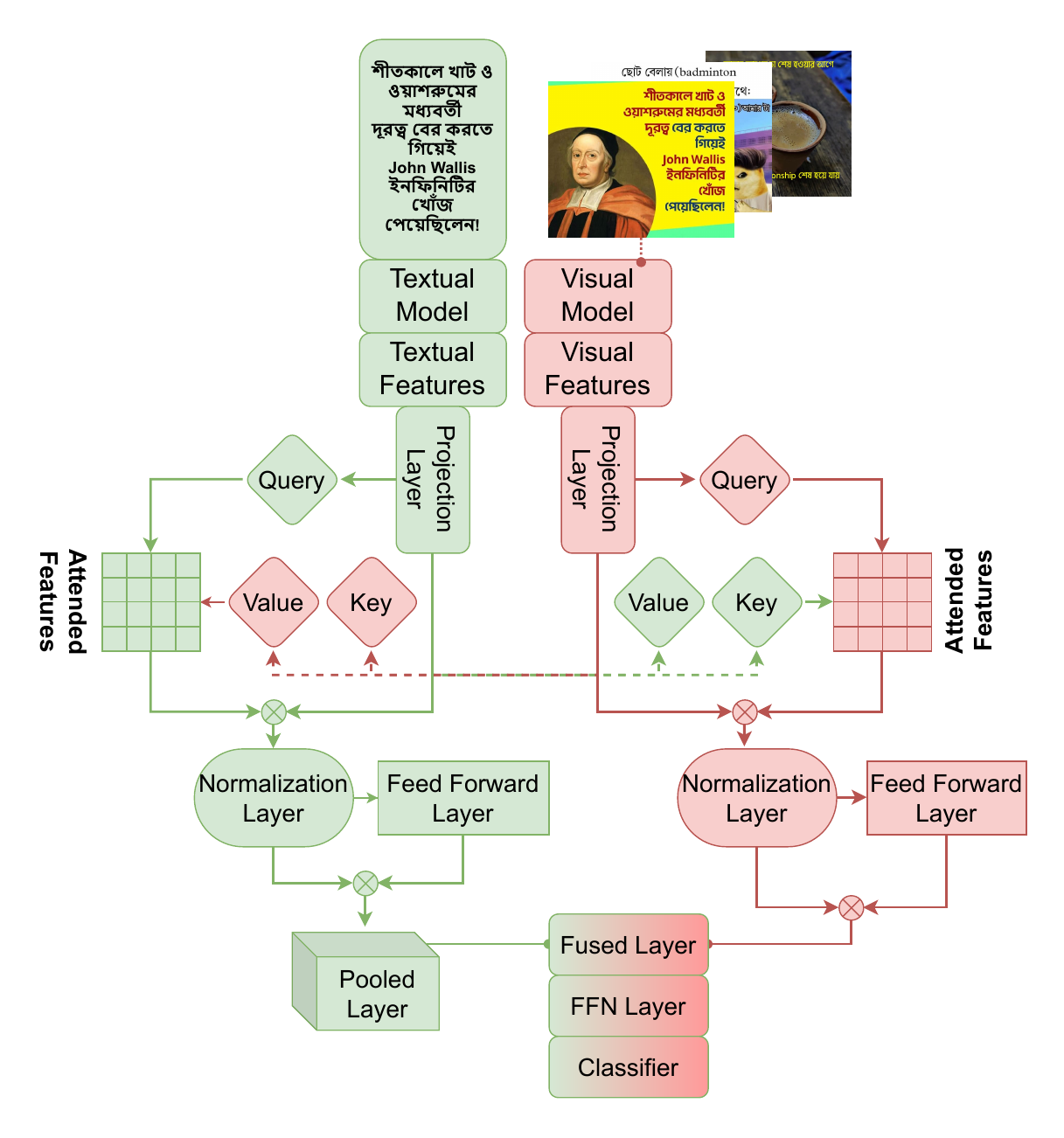}
    \caption{Proposed Multi-Modal Co-Attention Fusion Model (MCFM)}
    \label{fig:mcfm}
\end{figure}

Annotation followed open coding of 1,000 randomly sampled memes, revealing 15 categories grouped into the three overarching categories above. Three trained annotators—fluent in Bengali and code-mixed dialects—used Label Studio\cite{Label-studio} for labeling, viewing both image and OCR text. After a pilot phase on 300 memes (initial Fleiss’ $\kappa$ = 0.65), iterative calibration raised agreement to 0.79. Disagreements were resolved through expert arbitration considering irony, symbolism, and contextual nuance.  

\begin{figure}
    \centering
    \includegraphics[width=\linewidth]{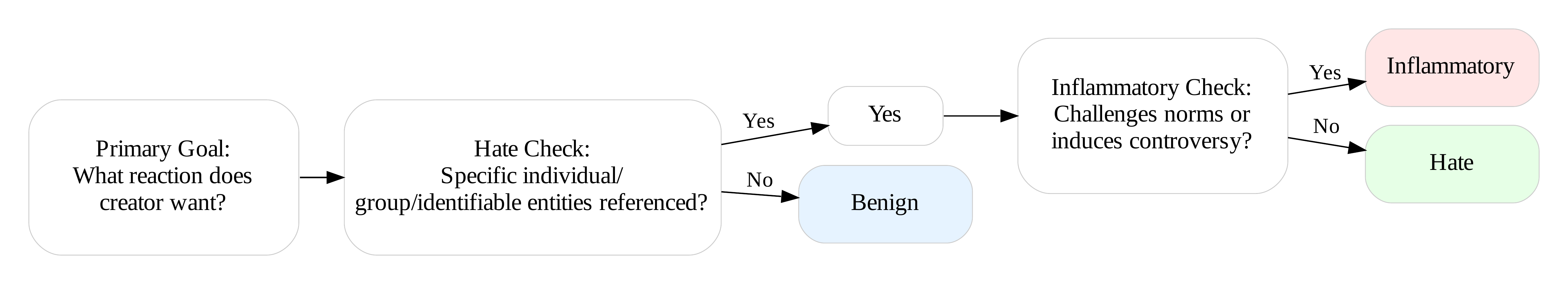}
    \caption{Annotation Decision Tree}\label{fig:Annotation Decision Tree}
\end{figure}

\begin{figure}
    \centering
    \includegraphics[width=\linewidth]{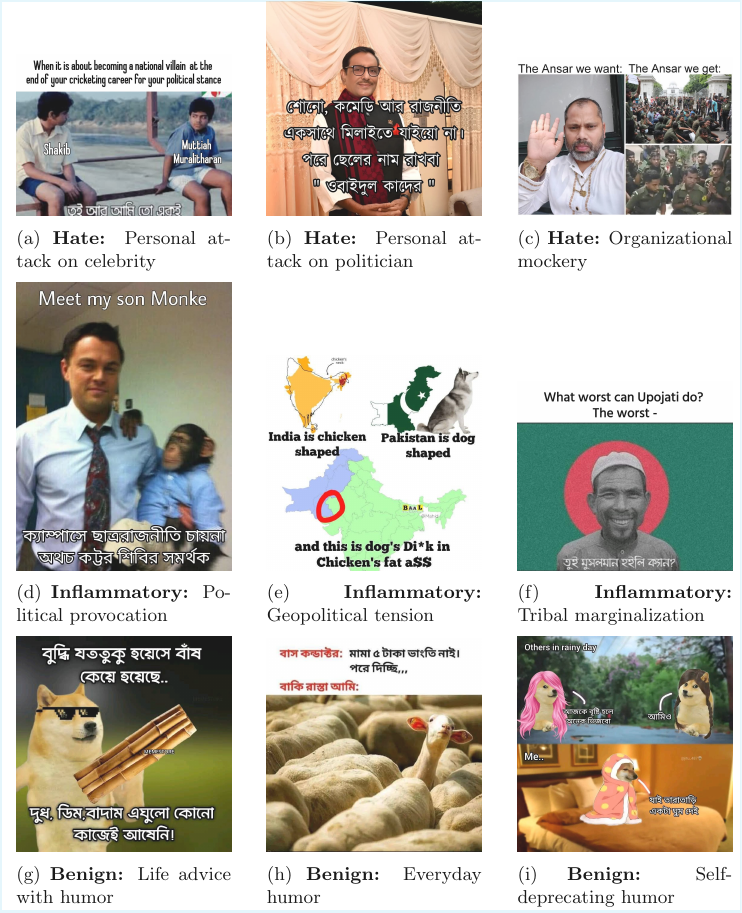}
    \caption{Examples Of Classes}
    \label{fig:example_of_classes}
\end{figure}

The final dataset comprises 3,247 annotated memes, divided into training (70\%), validation (15\%), and test (15\%) splits. Table~\ref{tab:Dataset Statistics: Class, Language, and Text Lengths} presents dataset statistics; Figs.~\ref{fig:Data Pre-Processing Diagram} and \ref{fig:Annotation Decision Tree} show the data collection and pre-processing workflow and decision rules for annotators. The dataset adheres to ethical, culturally sensitive, and bias-minimized annotation principles as \cite{ahsan2024}. Annotation inclusion and exclusion criteria are summarized in Tab.~\ref{tab:annotation-guidelines}. Figure \ref{fig:example_of_classes} illustrates examples from each category.

\begin{table}
\centering
\small
\caption{Dataset Statistics}
\resizebox{\columnwidth}{!}{
\begin{tabular}{lccc|ccc|c}
\toprule
\textbf{Category} & \multicolumn{3}{c|}{\textbf{Class Split}} & \multicolumn{3}{c|}{\textbf{Language Split}} & \textbf{Text Length (Words)} \\
\cmidrule(lr){2-4} \cmidrule(lr){5-7}
 & Train & Val & Test & Train & Val & Test & Mean / Max / Min / Mode / Med. \\
\midrule
Hate (HM)         & 811 & 174 & 173 & -- & -- & -- & -- \\
Inflammatory (IM) & 773 & 166 & 167 & -- & -- & -- & -- \\
Benign (BM)       & 688 & 147 & 148 & -- & -- & -- & -- \\
Bengali           & -- & -- & -- & 1190 & 255 & 256 & -- \\
Code-switched     & -- & -- & -- & 332  & 72  & 71  & -- \\
Code-mixed        & -- & -- & -- & 750  & 160 & 161 & -- \\
\midrule
\multicolumn{7}{r|}{\textbf{Overall}} & 121 / 15089 / 4 / 60 / 90 \\
\bottomrule
\end{tabular}}
\label{tab:Dataset Statistics: Class, Language, and Text Lengths}
\end{table}

\section{Methods}
\subsection{Models And Fusion Strategies}

\begin{table}
\centering
\caption{Textual, Visual, and Multi-Modal fusion Models}
\label{tab:models}
\setlength{\tabcolsep}{4pt} 
\renewcommand{\arraystretch}{0.9} 
\begin{tabular}{lll}
\toprule
\textbf{Category} & \textbf{Model / Strategy} & \textbf{Ref.} \\
\midrule
\multirow{5}{*}{Textual} 
& Banglish-BERT & \cite{banglishbert} \\
& mBERT & \cite{mbert} \\
& XLM-RoBERTa & \cite{xlm-roberta} \\
& MuRIL & \cite{muril} \\
& mDeBERTa-v3 & \cite{mDeBERT-1,mDeBERT-2} \\
\midrule
\multirow{6}{*}{Visual} 
& ResNet-50 & \cite{resnet50} \\
& ConvNeXt & \cite{convnext} \\
& ViT-B/16 & \cite{vit-1,vit-2} \\
& Swin V2 & \cite{swin} \\
& EfficientNet-B3 & \cite{efficientnet-b3} \\
& FocalNet-Base-LRF & \cite{focal-net} \\
\midrule
\multirow{5}{*}{Fusion} 
& Early Fusion &  \\
& Late Fusion &  \\
& CLIP & \cite{clip} \\
& Cross-Attention & \cite{cross-attention} \\
& MCFM(ours) &  \\
\bottomrule
\end{tabular}
\end{table}
Table~\ref{tab:models} summarizes the textual, visual, and multi-modal fusion strategies used. Textual encoders capture linguistic features in code-mixed or multilingual text, while visual encoders extract semantic and spatial features from meme images. Multi-modal fusion strategies—including early/late fusion, CLIP\cite{clip}-based integration, and attention-driven fusion enhance joint understanding of text and images. Fig~\ref{fig:mcfm} shows the architecture of our proposed model.

\subsection{Experimental Details}
To ensure reproducibility, the experimental details for all models—textual, visual, and multimodal—are consolidated in Table~\ref{tab:implementation details}. These specifications cover architectural variants, hyperparameters, and training procedures.

\subsection{Evaluation Metrics}
We evaluate performance on 3247 instances across three classes using Accuracy and Macro F1-score, prioritizing the latter due to minor class imbalance.  

\textbf{Accuracy} measures the proportion of correct predictions:
\begin{equation}
\text{Accuracy} = \frac{1}{3247} \sum_{i=1}^{3247} \mathbb{1}(y_i = \hat{y}_i),
\end{equation}
where \(y_i\) and \(\hat{y}_i\) are the true and predicted labels, and \(\mathbb{1}(\cdot)\) is the indicator function.  

\textbf{Macro F1-score} is the unweighted average of per-class F1-scores:
\begin{equation}
\text{F1}_{\text{macro}} = \frac{1}{3} \sum_{k=1}^{3} \frac{2 \cdot \text{Precision}_k \cdot \text{Recall}_k}{\text{Precision}_k + \text{Recall}_k},
\end{equation}
with \(\text{Precision}_k = \frac{TP_k}{TP_k + FP_k}\) and \(\text{Recall}_k = \frac{TP_k}{TP_k + FN_k}\), where \(TP_k\), \(FP_k\), and \(FN_k\) are true positives, false positives, and false negatives for class \(k\).

\section{Results}

\begin{table}[t]
\centering
\scriptsize
\setlength{\tabcolsep}{3.5pt} 
\caption{Model Performance (Accuracy / Macro F1)}
\label{tab:comparison-model-performance}
\begin{tabular}{@{}lcc|lcc@{}}
\toprule
\multicolumn{3}{c|}{\textbf{Textual Models}} & 
\multicolumn{3}{c}{\textbf{Visual Models}} \\ 
\cmidrule(lr){1-3} \cmidrule(lr){4-6}
Model & Acc. & F1 & Model & Acc. & F1 \\ 
\midrule
Banglish BERT & 0.7070 & 0.7109 & ResNet-50 & 0.5553 & 0.5509 \\
XLM-R & 0.7152 & 0.7162 & EffNet-B3 & 0.5205 & 0.5205 \\
M-BERT & 0.6865 & 0.6906 & ViT-B16 & 0.4734 & 0.4790 \\
Muril & 0.6475 & 0.6462 & ConvNext & 0.5738 & 0.5731 \\
\textit{MDeBERTa-V3} & \textbf{0.7193} & \textbf{0.7235} & \textit{Swin-V2} & \textbf{0.5922} & \textbf{0.5909} \\
FocalNet-lrf & 0.5512 & 0.5506 & & & \\
\midrule
\multicolumn{6}{c}{\textbf{Multimodal Models}} \\
Late Fusion & 0.6230 & 0.6201 & CLIP & 0.7295 & 0.7322 \\
Early Fusion & 0.6373 & 0.6440 & \textbf{MCFM} & \textbf{0.7746} & \textbf{0.7765} \\
Cross-Att (T$\rightarrow$I) & 0.7643 & 0.7649 & Cross-Att (I$\rightarrow$T) & 0.7684 & 0.7718 \\
\midrule
\multicolumn{6}{c}{\textbf{Improvement vs MDeBERTa-V3: +0.0553 Acc., +0.0530 F1}} \\
\bottomrule
\end{tabular}
\end{table}

\begin{figure}
    \centering
    \includegraphics[width=\linewidth]{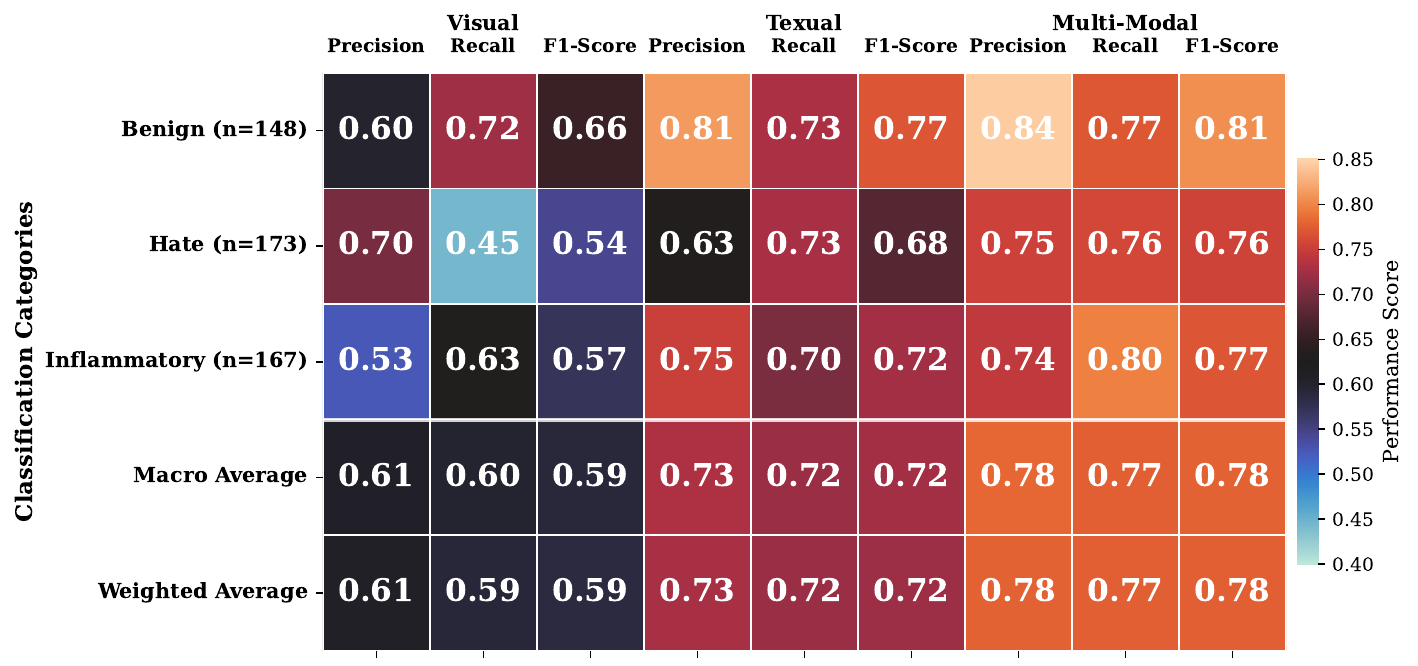}
    \caption{Class-Wise Performance Of Top Models}
    \label{fig:classification_heatmaps}
\end{figure}

Independent textual and visual classifiers show Table~\ref{tab:comparison-model-performance} that MDeBERTa-V3\cite{mDeBERT-1,mDeBERT-2} achieved the highest textual Macro F1 (0.7235), followed by XLM-RoBERTa\cite{xlm-roberta} (0.7162) and Banglish- BERT\cite{banglishbert} (0.7109), benefiting from disentangled attention and multilingual masked decoding. Visual-only models lagged: Swin Transformer-V2\cite{swin} (0.5909), ConvNeXt\cite{convnext} (0.5731), ResNet-50\cite{resnet50} (0.5509), reflecting low-res and blurry image challenge. Our fusion models leverage the best-performing unimodal models. Fusion models improved performance Table~\ref{tab:comparison-model-performance}: Early Fusion (0.6440) slightly outperformed Late Fusion, CLIP\cite{clip} exceeded textual baselines (0.7322), and our MCFM topped all (0.7765) via bidirectional modality alignment; Cross-Attention\cite{cross-attention} Image→Text (0.7718) indicates visual context aids semantic alignment. Multimodal models also reduced class-wise variance (see Fig~\ref{fig:classification_heatmaps}).

\subsection{Error Analysis}
\begin{figure}[ht]
    \centering
    \begin{subfigure}[b]{0.35\columnwidth}
        \centering
        \includegraphics[width=\linewidth]{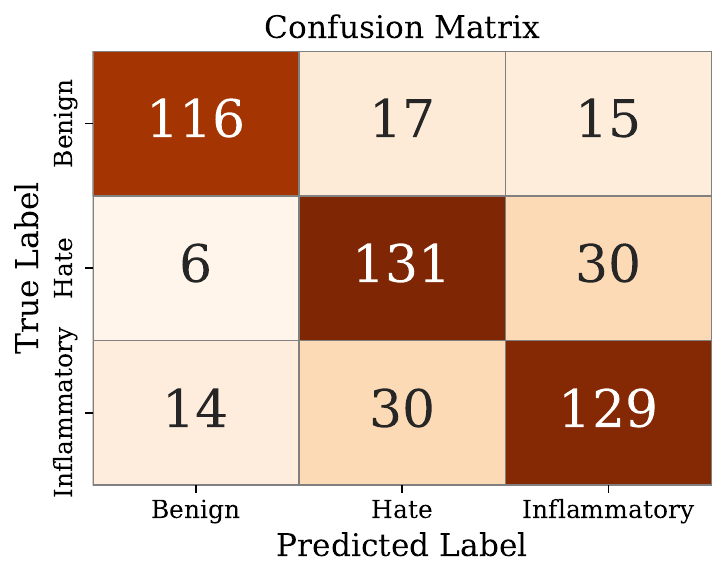}
        \caption{Confusion Matrix}
        \label{fig:confusion-matrix}
    \end{subfigure}
    \hfill
    \begin{subfigure}[b]{0.60\columnwidth}
        \centering
        \includegraphics[width=\linewidth]{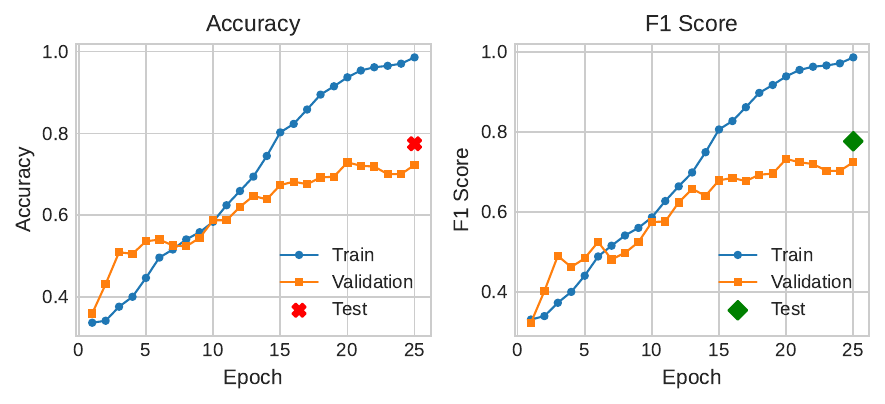}
        \caption{Performance Curves}
        \label{fig:performance-curves}
    \end{subfigure}
    \caption{Error analysis: confusion matrix and performance trends.}
    \label{fig:error-analysis}
\end{figure}

As shown in Fig.~\ref{fig:performance-curves}, the model overfit on the validation set despite fair test generalization. Fig.~\ref{fig:confusion-matrix} shows that Benign memes had the fewest errors (32: 17 hate, 15 Inflammatory), while Inflammatory memes had the most (44: 30 hate, 14 Benign). The hate class was often confused with Inflammatory (30 cases), reflecting semantic overlap.
High-confidence ($>$80\%) errors were dominated by the hate class (53.84\%), whereas low-confidence ($<$50\%) correct predictions indicated ambiguity between hate and Inflammatory categories.
Frequent errors stemmed from (i) conflicting text–image cues, (ii) unrecognized satire, sarcasm, or cultural references, and (iii) linguistic noise such as Bengali neologisms, OCR artifacts, and code-mixed text. Limited visual detail (e.g., small logos or symbols) further degraded accuracy.
Overall, these results highlight key multimodal moderation challenges—implicit humor, cultural nuance, and fine-grained semantic fusion—suggesting future focus on improved contextual and cross-modal understanding.

\begin{table*}[ht]
\centering
\caption{Experimental Details of Textual, Visual and Multi-Modal Models}
\label{tab:implementation details}
\scriptsize
\setlength{\tabcolsep}{4pt}
\begin{tabular}{p{2.5cm}|p{2.9cm}|p{2.5cm}|p{3.5cm}}
\hline
\textbf{Component} & \textbf{Textual (MDeBERTa-V3)} & \textbf{Visual (Swin-B)} & \textbf{Multi-Modal (Fusion)} \\
\hline
\hline
Base Model & \texttt{microsoft/} \texttt{mdeberta-v3-base} & Swin-B (\texttt{IMAGENET1K\_V1}) & Transformer + ImageNet-pretrained CNN \\
\hline
Hidden Size & 768 & -- & Text: 1024 \\
\hline
Input Size & Max token length = 256 & $224 \times 224$ & $224 \times 224$ \\
\hline
Tokenization / Preprocessing & Native tokenizer & Flip, rotation, jitter (train); resize + normalize (eval) & Tokenizer + image preprocessing \\
\hline
Special Tokens & [CLS], [SEP], [PAD] & -- & -- \\
\hline
Classification Head & ReLU $\rightarrow$ Dropout(0.5) $\rightarrow$ Linear (3 classes) & Dropout(0.5) $\rightarrow$ Linear (3 classes) & Dropout(0.5) $\rightarrow$ Linear \\
\hline
Fusion Architecture & -- & -- & Co-attention (MCFM), Cross-attention, CLIP-style \\
\hline
Fusion MLP & -- & -- & 2-layer MLP with GELU \\
\hline
Attention Heads & -- & -- & 8 \\
\hline
Optimizer & AdamW & Adam & AdamW \\
\hline
Learning Rate & $1 \times 10^{-5}$ & $1 \times 10^{-4}$ & $2 \times 10^{-5}$ \\
\hline
Weight Decay & 0.01 & -- & 0.01 \\
\hline
Batch Size & 8 & 8 & Effective 8 (grad acc = 2) \\
\hline
Max Epochs & 500 & 200 & -- \\
\hline
Early Stopping & Patience = 3 epochs & Patience = 3 epochs & Patience = 5 epochs (F1-based) \\
\hline
Loss Function & Cross-EntropyLoss & Cross-EntropyLoss & Focal Loss ($\alpha=1.0$, $\gamma=2.0$), Label Smoothing ($\epsilon=0.1$) \\
\hline
Class Balancing & -- & -- & Inverse frequency weighting \\
\hline
Gradient Clipping & Max norm = 1.0 & -- & Max norm = 1.0 \\
\hline
Learning Schedule & Linear warmup (10\%) & -- & Linear warmup (10\%) \\
\hline
Precision & Mixed (FP16/FP32) & AMP & Mixed Precision (FP16) \\
\hline
Gradient Accumulation & -- & -- & Steps = 2 \\
\hline
Efficiency Features & -- & -- & Gradient checkpointing, AMP \\
\hline
Regularization & Dropout, L2 decay & Dropout & Dropout, Label smoothing \\
\hline
Reproducibility & Seed 42, deterministic flags & Same & Same \\
\hline
Framework & PyTorch 1.12+, NVIDIA T4 & PyTorch 1.12+, NVIDIA T4 & PyTorch 1.12 + Transformers 4.20, NVIDIA T4 \\
\hline
\end{tabular}
\end{table*}

\section{Conclusion}
This paper presented Bn-HIB, a novel multimodal dataset of 3,247 memes across three classes, and proposed the Multimodal Co-Attention Fusion Model (MCFM). Experiments showed that MCFM surpasses fifteen state-of-the-art unimodal and multimodal baselines, validating its effectiveness. Future directions include expanding Bn-HIB to more domains and languages, enabling fine-grained categorization for richer analysis, Develop more lightweight version of MCFM and integrating  XAI to foster transparency, accountability, and user trust in content moderation.

\bibliographystyle{IEEEtran}
\bibliography{sn-bibliography}

@inproceedings{pramanick-etal-2021-momenta-multimodal,
  title        = {{MOMENTA}: A Multimodal Framework for Detecting Harmful Memes and Their Targets},
  author       = {Pramanick, Shraman and Sharma, Shivam and Dimitrov, Dimitar and Akhtar, Md. Shad and Nakov, Preslav and Chakraborty, Tanmoy},
  booktitle    = {Findings of the Association for Computational Linguistics: EMNLP 2021},
  pages        = {4439--4455},
  year         = {2021},
  address      = {Punta Cana, Dominican Republic},
  publisher    = {Association for Computational Linguistics},
}

@inproceedings{hossain-etal-2022-mute,
    title = "{MUTE}: A Multimodal Dataset for Detecting Hateful Memes",
    author = "Hossain, Eftekhar  and
      Sharif, Omar  and
      Hoque, Mohammed Moshiul",
    editor = "Hanqi, Yan  and
      Zonghan, Yang  and
      Ruder, Sebastian  and
      Xiaojun, Wan",
    booktitle = "Proceedings of the 2nd Conference of the Asia-Pacific Chapter of the Association for Computational Linguistics and the 12th International Joint Conference on Natural Language Processing: Student Research Workshop",
    month = nov,
    year = "2022",
    address = "Online",
    publisher = "Association for Computational Linguistics",
    url = "https://aclanthology.org/2022.aacl-srw.5/",
    doi = "10.18653/v1/2022.aacl-srw.5",
    pages = "32--39"
}

@misc{das2023banglaabusememedatasetbengaliabusive,
      title={BanglaAbuseMeme: A Dataset for Bengali Abusive Meme Classification}, 
      author={Mithun Das and Animesh Mukherjee},
      year={2023},
      eprint={2310.11748},
      archivePrefix={arXiv},
      primaryClass={cs.CV},
      url={https://arxiv.org/abs/2310.11748}, 
}

@InProceedings{10.1007/978-3-031-33231-9_21,
author="Karim, Md. Rezaul
and Dey, Sumon Kanti
and Islam, Tanhim
and Shajalal, Md
and Chakravarthi, Bharathi Raja",
editor="M, Anand Kumar
and Chakravarthi, Bharathi Raja
and B, Bharathi
and O'Riordan, Colm
and Murthy, Hema
and Durairaj, Thenmozhi
and Mandl, Thomas",
title="Multimodal Hate Speech Detection from Bengali Memes and Texts",
booktitle="Speech and Language Technologies for Low-Resource Languages ",
year="2023",
publisher="Springer International Publishing",
address="Cham",
pages="293--308",
isbn="978-3-031-33231-9"
}

@inproceedings{das-mukherjee-2023-banglaabusememe,
  title     = {{B}angla{A}buse{M}eme: A Dataset for {B}engali Abusive Meme Classification},
  author    = {Das, Mithun and Mukherjee, Animesh},
  booktitle = {Proceedings of the 2023 Conference on Empirical Methods in Natural Language Processing},
  pages     = {15498--15512},
  year      = {2023},
  address   = {Singapore},
  publisher = {Association for Computational Linguistics},
}

@inproceedings{ahsan2024,
  title={A Multimodal Framework to Detect Target Aware Aggression in Memes},
  author={Ahsan, Shawly and Hossain, Eftekhar and Sharif, Omar and Das, Avishek and Hoque, Mohammed Moshiul and Dewan, M. Ali Akber},
  booktitle={Proceedings of the 18th Conference of the European Chapter of the Association for Computational Linguistics Volume 1: Long Papers},
  pages={2487--2500},
  year={2024},
  organization={Association for Computational Linguistics}
}

@inproceedings{karim2022,
  title={Multimodal Hate Speech Detection from Bengali Memes and Texts},
  author={Karim, Md. Rezaul and Dey, Sumon Kanti and Islam, Tanhim and Shajalal, Md and Chakravarthi, Bharathi Raja},
  booktitle={International Conference on Speech and Language Technologies for Low-resource Languages},
  pages={293--308},
  year={2022},
  organization={Springer}
}

@inproceedings{suryawanshi2020a,
  title={Multimodal Meme Dataset (MultiOFF) for Identifying Offensive Content in Image and Text},
  author={Suryawanshi, Shardul and Chakravarthi, Bharathi Raja and Arcan, Mihael and Buitelaar, Paul},
  booktitle={Proceedings of the Second Workshop on Trolling, Aggression and Cyberbullying},
  pages={32--41},
  year={2020},
  organization={European Language Resources Association (ELRA)}
}

@inproceedings{hossain2024,
  title={Deciphering Hate: Identifying Hateful Memes and Their Targets},
  author={Hossain, Eftekhar and Sharif, Omar and Hoque, Mohammed Moshiul and Preum, Sarah M},
  booktitle={Proceedings of the 62nd Annual Meeting of the Association for Computational Linguistics (Volume 1: Long Papers)},
  pages={8347--8359},
  year={2024}
}

@inproceedings{banglishbert,
    title = "Not Low-Resource Anymore: Aligner Ensembling, Batch Filtering, and New Datasets for {B}engali-{E}nglish Machine Translation",
    author = "Hasan, Tahmid  and
      Bhattacharjee, Abhik  and
      Samin, Kazi  and
      Hasan, Masum  and
      Basak, Madhusudan  and
      Rahman, M. Sohel  and
      Shahriyar, Rifat",
    booktitle = "Proceedings of the 2020 Conference on Empirical Methods in Natural Language Processing (EMNLP)",
    month = nov,
    year = "2020",
    address = "Online",
    publisher = "Association for Computational Linguistics",
    url = "https://www.aclweb.org/anthology/2020.emnlp-main.207",
    doi = "10.18653/v1/2020.emnlp-main.207",
    pages = "2612--2623",
}

@article{mbert,
  author    = {Jacob Devlin and
               Ming{-}Wei Chang and
               Kenton Lee and
               Kristina Toutanova},
  title     = {{BERT:} Pre-training of Deep Bidirectional Transformers for Language
               Understanding},
  journal   = {CoRR},
  volume    = {abs/1810.04805},
  year      = {2018},
  url       = {http://arxiv.org/abs/1810.04805},
  archivePrefix = {arXiv},
  eprint    = {1810.04805},
  bibsource = {dblp computer science bibliography, https://dblp.org}
}

@article{xlm-roberta,
  author    = {Alexis Conneau and
               Kartikay Khandelwal and
               Naman Goyal and
               Vishrav Chaudhary and
               Guillaume Wenzek and
               Francisco Guzm{\'{a}}n and
               Edouard Grave and
               Myle Ott and
               Luke Zettlemoyer and
               Veselin Stoyanov},
  title     = {Unsupervised Cross-lingual Representation Learning at Scale},
  journal   = {CoRR},
  volume    = {abs/1911.02116},
  year      = {2019},
  url       = {http://arxiv.org/abs/1911.02116},
  eprinttype = {arXiv},
  eprint    = {1911.02116},
  bibsource = {dblp computer science bibliography, https://dblp.org}
}

@misc{muril,
      title={MuRIL: Multilingual Representations for Indian Languages},
      author={Simran Khanuja and Diksha Bansal and Sarvesh Mehtani and Savya Khosla and Atreyee Dey and Balaji Gopalan and Dilip Kumar Margam and Pooja Aggarwal and Rajiv Teja Nagipogu and Shachi Dave and Shruti Gupta and Subhash Chandra Bose Gali and Vish Subramanian and Partha Talukdar},
      year={2021},
      eprint={2103.10730},
      archivePrefix={arXiv},
      primaryClass={cs.CL}
}

@misc{mDeBERT-1,
      title={DeBERTaV3: Improving DeBERTa using ELECTRA-Style Pre-Training with Gradient-Disentangled Embedding Sharing}, 
      author={Pengcheng He and Jianfeng Gao and Weizhu Chen},
      year={2021},
      eprint={2111.09543},
      archivePrefix={arXiv},
      primaryClass={cs.CL}
}

@inproceedings{mDeBERT-2
,
title={DEBERTA: DECODING-ENHANCED BERT WITH DISENTANGLED ATTENTION},
author={Pengcheng He and Xiaodong Liu and Jianfeng Gao and Weizhu Chen},
booktitle={International Conference on Learning Representations},
year={2021},
url={https://openreview.net/forum?id=XPZIaotutsD}
}

@inproceedings{resnet50,
  title={Deep residual learning for image recognition},
  author={He, Kaiming and Zhang, Xiangyu and Ren, Shaoqing and Sun, Jian},
  booktitle={Proceedings of the IEEE conference on computer vision and pattern recognition},
  pages={770--778},
  year={2016}
}

@article{convnext,
  author    = {Zhuang Liu and
               Hanzi Mao and
               Chao{-}Yuan Wu and
               Christoph Feichtenhofer and
               Trevor Darrell and
               Saining Xie},
  title     = {A ConvNet for the 2020s},
  journal   = {CoRR},
  volume    = {abs/2201.03545},
  year      = {2022},
  url       = {https://arxiv.org/abs/2201.03545},
  eprinttype = {arXiv},
  eprint    = {2201.03545},
  bibsource = {dblp computer science bibliography, https://dblp.org}
}

@misc{vit-1,
      title={Visual Transformers: Token-based Image Representation and Processing for Computer Vision}, 
      author={Bichen Wu and Chenfeng Xu and Xiaoliang Dai and Alvin Wan and Peizhao Zhang and Zhicheng Yan and Masayoshi Tomizuka and Joseph Gonzalez and Kurt Keutzer and Peter Vajda},
      year={2020},
      eprint={2006.03677},
      archivePrefix={arXiv},
      primaryClass={cs.CV}
}

@inproceedings{vit-2,
  title={Imagenet: A large-scale hierarchical image database},
  author={Deng, Jia and Dong, Wei and Socher, Richard and Li, Li-Jia and Li, Kai and Fei-Fei, Li},
  booktitle={2009 IEEE conference on computer vision and pattern recognition},
  pages={248--255},
  year={2009},
  organization={Ieee}
}

@article{swin,
  author    = {Ze Liu and
               Han Hu and
               Yutong Lin and
               Zhuliang Yao and
               Zhenda Xie and
               Yixuan Wei and
               Jia Ning and
               Yue Cao and
               Zheng Zhang and
               Li Dong and
               Furu Wei and
               Baining Guo},
  title     = {Swin Transformer {V2:} Scaling Up Capacity and Resolution},
  journal   = {CoRR},
  volume    = {abs/2111.09883},
  year      = {2021},
  url       = {https://arxiv.org/abs/2111.09883},
  eprinttype = {arXiv},
  eprint    = {2111.09883},
  bibsource = {dblp computer science bibliography, https://dblp.org}
}

@article{efficientnet-b3,
  title={EfficientNet: Rethinking Model Scaling for Convolutional Neural Networks},
  author={Mingxing Tan and Quoc V. Le},
  journal={ArXiv},
  year={2019},
  volume={abs/1905.11946}
}

@misc{clip,
      title={Learning Transferable Visual Models From Natural Language Supervision}, 
      author={Alec Radford and Jong Wook Kim and Chris Hallacy and Aditya Ramesh and Gabriel Goh and Sandhini Agarwal and Girish Sastry and Amanda Askell and Pamela Mishkin and Jack Clark and Gretchen Krueger and Ilya Sutskever},
      year={2021},
      eprint={2103.00020},
      archivePrefix={arXiv},
      primaryClass={cs.CV},
      url={https://arxiv.org/abs/2103.00020}, 
}

@misc{focal-net,
      title={Focal Modulation Networks}, 
      author={Jianwei Yang and Chunyuan Li and Xiyang Dai and Jianfeng Gao},
      journal={Advances in Neural Information Processing Systems (NeurIPS)},
      year={2022}
}

@misc{cross-attention,
      title={Attention Is All You Need}, 
      author={Ashish Vaswani and Noam Shazeer and Niki Parmar and Jakob Uszkoreit and Llion Jones and Aidan N. Gomez and Lukasz Kaiser and Illia Polosukhin},
      year={2023},
      eprint={1706.03762},
      archivePrefix={arXiv},
      primaryClass={cs.CL},
      url={https://arxiv.org/abs/1706.03762}, 
}

@misc{click-mate,
  title        = {Auto Clicker Assistant:Your Ultimate Auto Tapper App},
  author       = {{INSCODE}},
  howpublished = {\url{https://play.google.com/store/apps/details?id=com.inscode.autoclicker}},
  note         = {Android app, updated December 10, 2024 (version as of listing) :contentReference[oaicite:1]{index=1}},
  year         = {2024},
  month        = dec,
}

@misc{Label-studio,
  title={{Label Studio}: Data labeling software},
  url={https://github.com/heartexlabs/label-studio},
  note={Open source software available from https://github.com/heartexlabs/label-studio},
  author={
    Maxim Tkachenko and
    Mikhail Malyuk and
    Andrey Holmanyuk and
    Nikolai Liubimov},
  year={2020-2022},
}

@article{kiela2020hateful,
  title={The hateful memes challenge: Detecting hate speech in multimodal memes},
  author={Kiela, Douwe and Firooz, Hamed and Mohan, Aravind and Goswami, Vedanuj and Singh, Amanpreet and Ringshia, Pratik and Testuggine, Davide},
  journal={Advances in neural information processing systems},
  volume={33},
  pages={2611--2624},
  year={2020}
}

@inproceedings{gomez2020exploring,
  title={Exploring Hate Speech Detection in Multimodal Publications},
  author={Gomez, Raul and Patel, Yash and Salvador, Amaia and et al.},
  booktitle={European Conference on Computer Vision},
  year={2020}
}

@inproceedings{debnath-etal-2025-exmute,
    title = "{E}x{M}ute: A Context-Enriched Multimodal Dataset for Hateful Memes",
    author = "Debnath, Riddhiman Swanan  and
      Firuj, Nahian Beente  and
      Shakib, Abdul Wadud  and
      Sultana, Sadia  and
      Islam, Md Saiful",
    editor = "Weerasinghe, Ruvan  and
      Anuradha, Isuri  and
      Sumanathilaka, Deshan",
    booktitle = "Proceedings of the First Workshop on Natural Language Processing for Indo-Aryan and Dravidian Languages",
    month = jan,
    year = "2025",
    address = "Abu Dhabi",
    publisher = "Association for Computational Linguistics",
    url = "https://aclanthology.org/2025.indonlp-1.10/",
    pages = "83--89"
}

@misc{mishra2023memotion3datasetsentiment,
      title={Memotion 3: Dataset on Sentiment and Emotion Analysis of Codemixed Hindi-English Memes}, 
      author={Shreyash Mishra and S Suryavardan and Parth Patwa and Megha Chakraborty and Anku Rani and Aishwarya Reganti and Aman Chadha and Amitava Das and Amit Sheth and Manoj Chinnakotla and Asif Ekbal and Srijan Kumar},
      year={2023},
      eprint={2303.09892},
      archivePrefix={arXiv},
      primaryClass={cs.CL},
      url={https://arxiv.org/abs/2303.09892}
}

@inproceedings{10.1007/978-3-031-28244-7_7,
author = {Bandyopadhyay, Dibyanayan and Kumari, Gitanjali and Ekbal, Asif and Pal, Santanu and Chatterjee, Arindam and BN, Vinutha},
title = {A Knowledge Infusion Based Multitasking System for Sarcasm Detection in Meme},
year = {2023},
booktitle = {Advances in Information Retrieval: ECIR 2023},
pages = {101--117},
publisher = {Springer-Verlag},
doi = {10.1007/978-3-031-28244-7_7}
}

@inproceedings{inproceedings,
author = {Zhou, Yi and Chen, Zhenhao and Yang, Huiyuan},
year = {2021},
title = {Multimodal Learning For Hateful Memes Detection},
booktitle = {ICMEW 2021},
pages = {1--6},
doi = {10.1109/ICMEW53276.2021.9455994}
}

@article{DBLP:journals/corr/abs-2005-04790,
  author       = {Douwe Kiela and Hamed Firooz and Aravind Mohan and Vedanuj Goswami and Amanpreet Singh and Pratik Ringshia and Davide Testuggine},
  title        = {The Hateful Memes Challenge: Detecting Hate Speech in Multimodal Memes},
  journal      = {CoRR},
  volume       = {abs/2005.04790},
  year         = {2020},
  url          = {https://arxiv.org/abs/2005.04790}
}

@article{SHANG2021102664,
title = {AOMD: An analogy-aware approach to offensive meme detection on social media},
journal = {Information Processing and  Management},
volume = {58},
number = {5},
pages = {102664},
year = {2021},
author = {Lanyu Shang and Yang Zhang and Yuheng Zha and Yingxi Chen and Christina Youn and Dong Wang},
doi = {https://doi.org/10.1016/j.ipm.2021.102664}
}

@inproceedings{suryawanshi-chakravarthi-2021-findings,
    title = "Findings of the Shared Task on Troll Meme Classification in {T}amil",
    author = "Suryawanshi, Shardul and Chakravarthi, Bharathi Raja",
    booktitle = "Proceedings of the First Workshop on Speech and Language Technologies for Dravidian Languages",
    year = "2021",
    pages = "126--132",
    url = "https://aclanthology.org/2021.dravidianlangtech-1.16/"
}

@inproceedings{kumari2023,
  author = {Kumari, Gitanjali and Bandyopadhyay, Dibyanayan and Ekbal, Asif and Pal, Santanu},
  title = {HateMeme-Hindi: A Multimodal Dataset for Hateful Meme Detection in Hindi},
  booktitle = {Proceedings of the 20th International Conference on Natural Language Processing (ICON)},
  year = {2023}
}

@inproceedings{manukonda-kodali-2025-bytesizedllm-dravidianlangtech-2025,
    title = "byte{S}ized{LLM}@{D}ravidian{L}ang{T}ech 2025: Multimodal Misogyny Meme Detection in Low-Resource {D}ravidian Languages Using Transliteration-Aware {XLM}-{R}o{BERT}a, {R}es{N}et-50, and Attention-{B}i{LSTM}",
    author = "Manukonda, Durga Prasad  and
      Kodali, Rohith Gowtham",
    booktitle = "Proceedings of the Fifth Workshop on Speech, Vision, and Language Technologies for Dravidian Languages",
    month = may,
    year = "2025",
    address = "Acoma, The Albuquerque Convention Center, Albuquerque, New Mexico",
    publisher = "Association for Computational Linguistics",
    url = "https://aclanthology.org/2025.dravidianlangtech-1.15/",
    doi = "10.18653/v1/2025.dravidianlangtech-1.15",
    pages = "86--91",
    ISBN = "979-8-89176-228-2"
}

@article{KUMARI2025101781,
title = {Identifying offensive memes in low-resource languages: A multi-modal multi-task approach using valence and arousal},
journal = {Computer Speech and  Language},
volume = {92},
pages = {101781},
year = {2025},
issn = {0885-2308},
doi = {https://doi.org/10.1016/j.csl.2025.101781},
url = {https://www.sciencedirect.com/science/article/pii/S0885230825000063},
author = {Gitanjali Kumari and Dibyanayan Bandyopadhyay and Asif Ekbal and Arindam Chatterjee and Vinutha B.N.}
}

@article{KAPIL2025100133,
title = {A transformer based multi task learning approach to multimodal hate speech detection},
journal = {Natural Language Processing Journal},
volume = {11},
pages = {100133},
year = {2025},
issn = {2949-7191},
doi = {https://doi.org/10.1016/j.nlp.2025.100133},
url = {https://www.sciencedirect.com/science/article/pii/S2949719125000093},
author = {Prashant Kapil and Asif Ekbal}
}

@inproceedings{el-sayed-nasr-2024-aast-nlp,
    title = "{AAST}-{NLP} at Multimodal Hate Speech Event Detection 2024 : A Multimodal Approach for Classification of Text-Embedded Images Based on {CLIP} and {BERT}-Based Models.",
    author = "El-Sayed, Ahmed  and
      Nasr, Omar",
    editor = {H{\"u}rriyeto{\u{g}}lu, Ali  and
      Tanev, Hristo  and
      Thapa, Surendrabikram  and
      Uludo{\u{g}}an, G{\"o}k{\c{c}}e},
    booktitle = "Proceedings of the 7th Workshop on Challenges and Applications of Automated Extraction of Socio-political Events from Text (CASE 2024)",
    month = mar,
    year = "2024",
    address = "St. Julians, Malta",
    publisher = "Association for Computational Linguistics",
    url = "https://aclanthology.org/2024.case-1.19/",
    pages = "139--144"
}

@misc{rajput2022hatenotdetectinghate,
      title={Hate Me Not: Detecting Hate Inducing Memes in Code Switched Languages}, 
      author={Kshitij Rajput and Raghav Kapoor and Kaushal Rai and Preeti Kaur},
      year={2022},
      eprint={2204.11356},
      archivePrefix={arXiv},
      primaryClass={cs.LG},
      url={https://arxiv.org/abs/2204.11356}, 
}

@misc{GeminiApi,
    author = {{Google}},
    note = {Generative AI chat api},
    year = {2025},
    month = {},
    day = {}
}

\end{document}